\documentclass{article}

\usepackage{arxiv}

\usepackage{subcaption}
\usepackage{amsmath}

\usepackage[utf8]{inputenc} 
\usepackage[T1]{fontenc}    
\usepackage{hyperref}       
\usepackage{url}            
\usepackage{booktabs}       
\usepackage{amsfonts}       
\usepackage{nicefrac}       
\usepackage{microtype}      
\usepackage{cleveref}       
\usepackage{graphicx}
\usepackage{natbib}
\usepackage{doi}

\title{Challenging the Black Box: A Comprehensive Evaluation of Attribution Maps of CNN Applications in Agriculture and Forestry}

\author{ \href{https://orcid.org/0000-0002-7523-5694}{\includegraphics[scale=0.06]{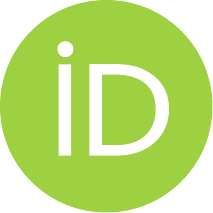}\hspace{1mm}Lars~Nieradzik}\\
	Image Processing Department\\
	Fraunhofer ITWM\\
	Fraunhofer Platz 1, 67663, Kaiserslautern\\
	\texttt{lars.nieradzik@itwm.fraunhofer.de}\\
	\And
	\href{https://orcid.org/0000-0002-9821-1636}{\includegraphics[scale=0.06]{orcid.pdf}\hspace{1mm}Henrike~Stephani} \\
	Image Processing Department\\
	Fraunhofer ITWM\\
	Fraunhofer Platz 1, 67663, Kaiserslautern\\
	\texttt{henrike.stephani@itwm.fraunhofer.de}\\
	\And
	\href{https://orcid.org/0009-0001-7547-269X}{\includegraphics[scale=0.06]{orcid.pdf}\hspace{1mm}Jördis ~Sieburg-Rockel} \\
	Thünen Institute of Wood Research\\
	Leuschnerstraße 91, 21031, Hamburg\\
	\texttt{joerdis.sieburg-rockel@thuenen.de} \\
	\And
	\href{https://orcid.org/0009-0009-6611-3140}{\includegraphics[scale=0.06]{orcid.pdf}\hspace{1mm}Stephanie~Helmling} \\
	Thünen Institute of Wood Research\\
	Leuschnerstraße 91, 21031, Hamburg\\
	\texttt{stephanie.helmling@thuenen.de} \\
	\And
	\href{https://orcid.org/0009-0007-2249-2797}{\includegraphics[scale=0.06]{orcid.pdf}\hspace{1mm}Andrea~Olbrich} \\
	Thünen Institute of Wood Research\\
	Leuschnerstraße 91, 21031, Hamburg\\
	\texttt{andrea.olbrich@thuenen.de} \\
	\And
	\href{https://orcid.org/0000-0002-1327-1243}{\includegraphics[scale=0.06]{orcid.pdf}\hspace{1mm}Janis~Keuper} \\
	Institute of Machine Learning and Analysis (IMLA)\\
	Offenburg University\\
	Badstr. 24, 77652, Offenburg\\
	\texttt{jkeuper@ad.hs-offenburg.de}\\
}

\renewcommand{\shorttitle}{Challenging the Black Box}

\hypersetup{
pdftitle={A template for the arxiv style},
pdfsubject={q-bio.NC, q-bio.QM},
pdfauthor={David S.~Hippocampus, Elias D.~Striatum},
pdfkeywords={First keyword, Second keyword, More},
}

\begin{document}
\maketitle

\begin{abstract}In this study, we explore the explainability of neural networks in agriculture and forestry, specifically in fertilizer treatment classification and wood identification. The opaque nature of these models, often considered 'black boxes', is addressed through an extensive evaluation of state-of-the-art Attribution Maps (AMs), also known as class activation maps (CAMs) or saliency maps. Our comprehensive qualitative and quantitative analysis of these AMs uncovers critical practical limitations. Findings reveal that AMs frequently fail to consistently highlight crucial features and often misalign with the features considered important by domain experts. These discrepancies raise substantial questions about the utility of AMs in understanding the decision-making process of neural networks. Our study provides critical insights into the trustworthiness and practicality of AMs within the agriculture and forestry sectors, thus facilitating a better understanding of neural networks in these application areas.
\end{abstract}


\section{Introduction}

The application of neural networks in agriculture and forestry has proven beneficial in various tasks such as wood identification \cite{nieradzik2023automating}, plant phenotyping, yield prediction, and disease detection. However, a significant roadblock to wider adoption is the inherently opaque nature of these models, which tends to dampen user confidence due to their limited explainability.

\begin{figure}[ht]
  \centering
  \includegraphics[scale=1.0]{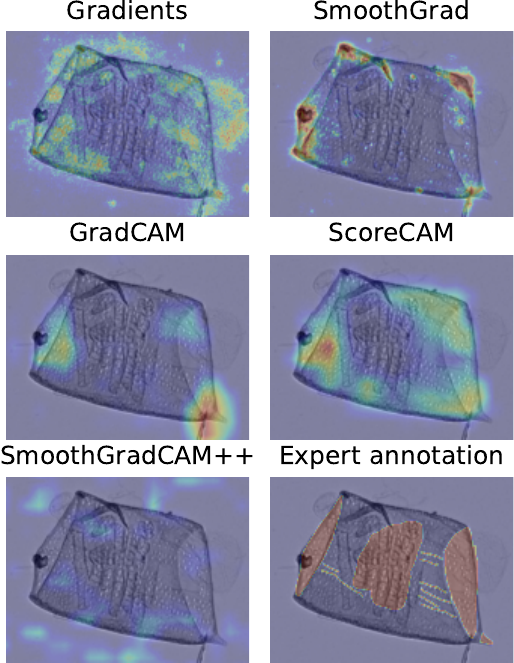}
  
   \caption{Visualization of different attribution maps (AM) on the same input image of a wood identification dataset. All the AMs focus on different regions that are also different from the expert annotation. Notably, SmoothGradCAM++ \cite{DBLP:journals/corr/abs-1908-01224} appears to exclusively show noise.}
   \label{fig:diffattrmap}
\end{figure}

Wood identification serves as a prime example of this challenge. It remains unclear whether the decisions of neural networks focus on the same set of features as a human expert would. Humans use 163 structural features defined by the International Association of Wood Anatomists \cite{wheeler1989iawa} for microscopic descriptions of approximately 8,700 timbers, as collected in various databases \cite{Jorgo, insidewood, openagrar_mods_00085331}.

To demystify these black-box models, Attribution Methods have emerged as standard tools for visualizing the decision processes in neural networks \cite{zhang2021survey}. These methods, particularly relevant in Computer Vision tasks with image inputs, compute Attribution Maps (AMs), also known as Saliency Maps, for individual images based on trained models. AMs aim to provide human-interpretable visualizations that reveal the weighted impact of image regions on model predictions, enabling intuitive explanations of the complex internal mappings. Despite their potential, the practical adoption of AMs in various domains remains limited.

Addressing this gap, our paper conducts a thorough evaluation of multiple state-of-the-art attribution maps using two real-world datasets from the agriculture and forestry domain. We have trained state-of-the-art Convolutional Neural Networks (CNNs) on a wood identification dataset and a dataset concerning fertilizer treatments for nutrient deficiencies in winter wheat and winter rye. The key contributions of this paper include:

\begin{itemize}
\item A comprehensive analysis of state-of-the-art attribution maps for wood identification and fertilizer treatment, both qualitatively and quantitatively.
\item We identify significant variance among attribution methods, leading to inconsistent region highlighting and excessive noise in certain methods.
\item Our results raise concerns regarding attribution maps that display excessive feature sharing across distinct classes, suggesting potential issues.
\item In collaboration with wood anatomists, key features were annotated in the wood identification dataset. These annotations were used to measure the alignment with respect to the attribution maps. Notably, none of the maps showed high alignment with expert annotations.
\end{itemize}

By investigating these aspects, this study provides valuable insights into the effectiveness and reliability of attribution maps for wood species identification and fertilizer treatment classification, benefiting domain experts in agriculture and forestry.

\section{Related Work}

\subsection{Attribution methods}

The field of attribution methods has witnessed significant growth in recent years, with numerous techniques being developed and widely used in the context of neural networks. Full back-propagation methods, such as Gradients \cite{DBLP:journals/corr/SimonyanVZ13}, have played a fundamental role in the early approaches to attribution maps for classification models. These methods compute gradients of a learned neural network with respect to a given input, providing insights into the importance of different image regions. DeConvNet \cite{DBLP:journals/corr/ZeilerF13} and Guided backpropagation \cite{DBLP:journals/corr/SpringenbergDBR14} were introduced as extensions to Gradients, modifying the gradients to allow the backward flow of negative gradients. Another variant, SmoothGrad \cite{DBLP:journals/corr/SmilkovTKVW17}, enhances the gradient computation by adding Gaussian noise to the input and averaging the results.

Path backpropagation methods take a different approach by parameterizing a path from a baseline image to the input image and computing derivatives along this path. Integrated Gradients \cite{DBLP:journals/corr/SundararajanTY17}, for example, uses a straight line path between a black (all-zero) image and the test input, and integrates the partial derivatives of the neural network along this path. Variations of Integrated Gradients, such as Blur Integrated Gradients \cite{DBLP:journals/corr/abs-2004-03383} and Guided Integrated Gradients \cite{DBLP:journals/corr/abs-2106-09788}, have been proposed to improve the path initialization and computation.

Class Activation Maps (CAMs) offer an alternative to computing gradients with respect to the input. CAM methods stop the back-propagation of gradients at a chosen layer of the network. GradCAM \cite{Selvaraju_2019}, a popular CAM approach, computes an attribution map by summing weighted activations at the chosen layer. GradCAM++ \cite{Chattopadhay_2018} further enhances the original GradCAM by incorporating different gradient weighting schemes. Smooth GradCAM++\cite{DBLP:journals/corr/abs-1908-01224} adds Gaussian noise to the input, similar to SmoothGrad, to improve the visualization. Other CAM variants, such as LayerCAM \cite{9462463} and XGradCAM \cite{DBLP:journals/corr/abs-2008-02312}, introduce different weighting schemes for the computed gradients.

In contrast to gradient-based methods, ScoreCAM \cite{wang2020scorecam} does not rely on gradient computations for attribution. Instead, it measures the importance of channels in an intermediate layer by observing the change in confidence when removing parts of the activation values.

Beyond these well-known approaches, numerous other attribution map methods have been proposed, building upon and combining existing techniques \cite{DBLP:journals/corr/abs-1711-06104}. These methods include SS-CAM \cite{https://doi.org/10.48550/arxiv.2006.14255}, IS-CAM \cite{DBLP:journals/corr/abs-2010-03023}, Ablation-CAM \cite{9093360}, FD-CAM \cite{https://doi.org/10.48550/arxiv.2206.08792}, Group-CAM \cite{DBLP:journals/corr/abs-2103-13859}, Poly-CAM \cite{https://doi.org/10.48550/arxiv.2204.13359}, Zoom-CAM \cite{DBLP:journals/corr/abs-2010-08644}, and EigenCAM \cite{DBLP:journals/corr/abs-2008-00299}, each introducing unique modifications and improvements to the attribution map generation process.

Additionally, black-box methods offer an alternative approach by masking the input in various ways. For instance, RISE \cite{https://doi.org/10.48550/arxiv.1806.07421} and its precursor, introduced in earlier works \cite{DBLP:journals/corr/ZeilerF13}, randomly occlude the input image and record the resulting change in class probabilities. Several other black-box methods have also been proposed in the literature \cite{DBLP:journals/corr/abs-1910-08485,DBLP:journals/corr/FongV17,DBLP:journals/corr/abs-1806-07421,DBLP:journals/corr/RibeiroSG16}.

Overall, the field of attribution methods offers a diverse range of techniques, each with its own strengths and limitations.

\subsection{Evaluation of attribution methods}

In this work, we aim to evaluate and compare various state-of-the-art attribution map methods in the context of wood species identification and fertilizer treatment classification, specifically focusing on their applicability and usefulness in real-world domains such as agriculture and forestry.

A similar study conducted by \cite{Saporta2022} in the medical field evaluated attribution maps for interpreting chest x-rays. However, their analysis focused mainly on comparing attribution maps with human annotations. In contrast, we extend our study to scenarios where annotations are not available. It is important to note that annotations are not necessarily the basis for the network's decision. As a surrogate, we measure the consistency between different attribution maps, use established metrics from the existing literature, and perform a thorough qualitative analysis.

In another relevant research endeavor in the area of plant phenotyping \cite{doi:10.34133/2019/9237136}, various methods for visualizing network behavior were considered. While qualitative analysis is certainly valuable, our approach places a greater emphasis on the quantitative aspects.

Finally, we argue for the use of modern CNN architectures. The aforementioned works relied on models such as InceptionV3 \cite{szegedy2015rethinking}, DenseNet121 \cite{huang2018densely}, or ResNet152 \cite{he2015deep}, which are considered somewhat outdated in light of the rapidly evolving field of Deep Learning. These models produce suboptimal results on both standard and real-world datasets \cite{fang2023does}. We propose to use more modern architectures that potentially produce better class activation maps \cite{DBLP:journals/corr/abs-2201-03545} (as seen in \cite{DBLP:journals/corr/abs-1905-11946}).

\section{Method}

\subsection{Consistency}

The output of attribution methods are matrices, normalized within the range of $[0, 1]^{n \times m}$, where $n$ and $m$ represent the image's dimensions. Given that these matrices do not typically exhibit high structural complexity necessitating adjustments for shifts or color adaptations, we opt for straightforward metrics performing pixel-wise comparisons between the saliency maps. A low similarity index among saliency maps indicates reduced consistency, as different regions are deemed important by different maps. Moreover, when all the saliency maps roughly agree with each other, it means that the choice of the saliency map is not that important. However, in situations where all saliency maps disagree, one may prove superior to the rest.

We introduce two metrics to measure consistency. The first metric is the Pearson correlation coefficient, defined as:
$$r_{xy}={\frac {\sum _{i=1}^{n}(x_{i}-{\bar {x}})(y_{i}-{\bar {y}})}{{\sqrt {\sum _{i=1}^{n}(x_{i}-{\bar {x}})^{2}}}{\sqrt {\sum _{i=1}^{n}(y_{i}-{\bar {y}})^{2}}}}}\,,$$

where $x_i$ is the $i$th pixel in the AM and $\bar{x}$ is the sample mean. Similarly, $y_i$ is the $i$th pixel of the second AM. The output range of $r_{xy}$ is $[-1, 1]$, with $1$ indicating the highest consistency and $\leq 0$ a low consistency. $r_{xy} = 0$ is intuitively random noise and $r_{xy} = -1$ an "inverted" saliency map.

The second metric is the Jensen–Shannon divergence (JSD). We assume that each pixel is the result of a binary regression model, which classified the pixel as either important or unimportant (Bernoulli distribution). Then we can compare the distribution $X$ of this pixel against a second distribution $Y$ to see how similar the two distributions are.

First, let us define the Kullback–Leibler divergence between all the pixels of two saliency maps.

$$D(X\parallel Y) = \frac{1}{n}\sum_{i=1}^n x_i\log_2\left(\frac{x_i}{y_i}\right) + (1 - x_i)\log_2\left(\frac{1 - x_i}{1 - y_i}\right)$$

The formula $D(X\parallel Y)$, while informative, lacks symmetry and does not have an upper bound of $1$, making the results less interpretable. To yield more understandable numbers, we opt for JSD instead. This is a smoothed and symmetric variant of the Kullback-Leibler divergence, thereby offering an enhanced interpretability. It is defined as

$$0 \leq {\rm{JSD}}(X\parallel Y)={\frac {1}{2}}D(X\parallel M)+{\frac {1}{2}}D(Y\parallel M) \leq 1\,,$$

where $M={\frac {1}{2}}(X+Y)$.

The results of both metrics can be visualized in a confusion matrix. We take the average of the upper triangular elements of the confusion matrix to obtain a single value. Then the consistency for Pearson's r is defined as

$$\text{Consistency}_{\text{Corr}} = \frac{2}{m(m-1)}\sum_{i=1}^{m-1}\sum_{j=i+1}^{m}r_{x_ix_j}$$

We define $\text{Consistency}_{\text{JSD}}$ in the same way.

\subsection{Qualitative and quantitative evaluation of saliency maps}

In situations where saliency maps demonstrate substantial inconsistency among themselves, the selection of the most suitable saliency map for the specific task becomes critical. Numerous metrics have been proposed to assess the quality of these saliency maps \cite{Chattopadhay_2018, DBLP:journals/corr/abs-2104-10252, DBLP:journals/corr/FongV17, DBLP:journals/corr/abs-1806-07421, DBLP:journals/corr/abs-2201-13291, DBLP:journals/corr/Zhang0BSS16, https://doi.org/10.48550/arxiv.2208.06175}, with the intent to discern which one is predicted to yield the best performance for a particular dataset, or even across datasets.

Among the state-of-the-art metrics, two prominent ones are "Insertion" and "Deletion". These metrics will be utilized in our comparison analysis to evaluate the saliency maps' performance and determine their effectiveness.

Deletion and Insertion were proposed by \cite{DBLP:journals/corr/FongV17} and \cite{DBLP:journals/corr/abs-1806-07421}. It is an iterative process of pixel deletion or insertion within the test image. The pixels are ordered by the importance given by the saliency map. For instance, the deletion process initiates with the input test image, and then sequentially masks the subsequent important regions with a value of $0$. Each modified image is given to the neural network to produce a probability. The first image is the original image and the last image is a black image. This process can be visualized by plotting the count of inserted or deleted pixels (x-axis) against the probability of the target class (y-axis). To summarize this plot into a scalar value, the area under the curve (AUC) is calculated.

\begin{figure*}[ht]
\begin{subfigure}{.5\textwidth}
  \centering
  \includegraphics[scale=0.42]{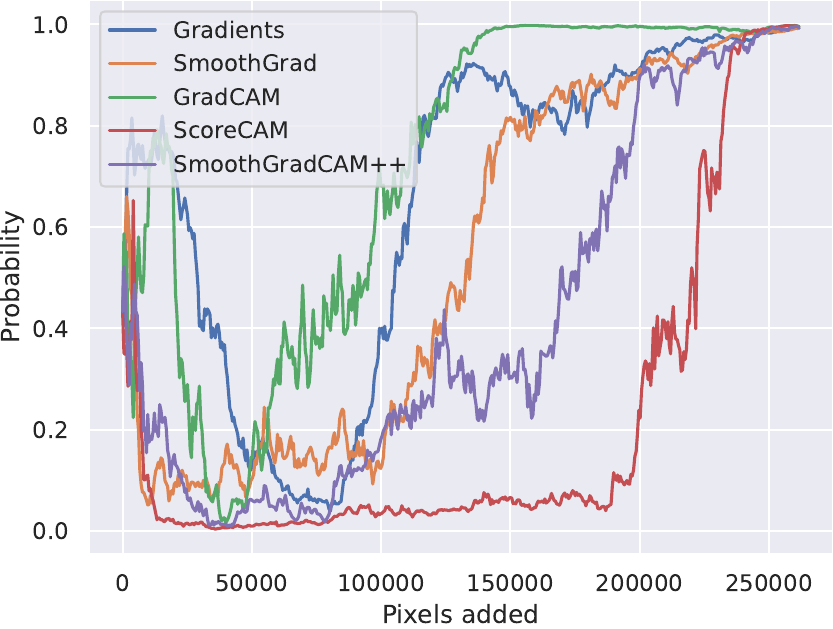}
  \caption{Insertion with blurring}
\end{subfigure}
\begin{subfigure}{.5\textwidth}
  \centering
  \includegraphics[scale=0.42]{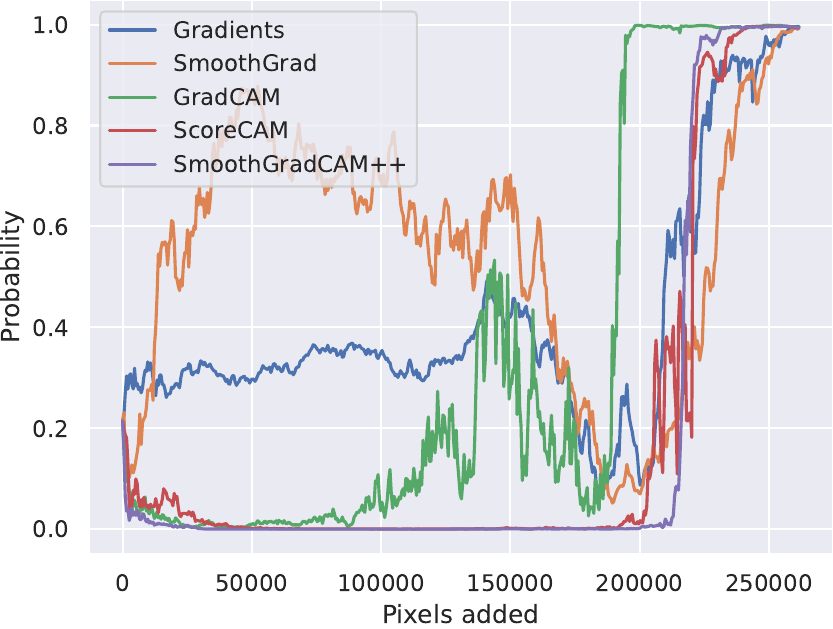}
  \caption{Insertion without blurring}
\end{subfigure}
\caption{The "Insertion" metric evaluated on an example image illustrates the impact of the parameter "blurring" in comparison to "no blurring". In (a), the starting image is a completely blurred image. In (b), the starting image is a black image. We input this modified image into the neural network to obtain a probability (see y-axis). First, the most important pixels of the original image are inserted. Then gradually less important pixels are inserted. At each step, the network predicts the probability of this modified image. The process ends with the complete original image and the original probability. The best saliency map in the plots is determined by computing the area under the curve (AUC).}
\label{fig:blurringandnot}
\end{figure*}

While these evaluation methods are sensible, they have certain intrinsic limitations. They are not comprehensive measures, as they overlook critical factors such as the noise level in the saliency map and the map's degree of user-informativeness. Moreover, these methods are influenced by their respective parameters. Frequently, they work based on concealing parts of the image that the neural network deems significant. The process of obscuring these areas, whether through blurring or masking, introduces another parameter. The size of the blurring/masking kernel or the number of pixels chosen for image modification at each iteration can have a significant impact on the metric's result.  An example can be seen in \cref{fig:blurringandnot}. Therefore, it is not enough to look only at the metrics to prove that a particular saliency map reflects the correct decision process of a neural network.

Thus, visual comparison of the results with expert annotations remains indispensable for verification. Instead of comparing the attribution map purely with previously mentioned metrics, we also propose a comparison with manual feature annotations. It is crucial to note that convolutional neural networks (CNNs) might select different image features for decision-making than a human expert would. Nonetheless, utilizing human annotations offers a form of "ground truth" that, while it may not represent "the" definitive decision-making process of the CNN, at least provides "a" valid process for this classification problem.

Finally, it is also worthwhile examining the same saliency map of different target classes. From the perspective of a human expert, the highlighting of a region in the saliency map signifies the presence of a particular feature. Therefore, if a region appears in the saliency maps of multiple classes, it suggests that those classes share a common feature. However, for a neural network, the decision-making process is not solely based on the existence or absence of specific regions in the saliency map. Even a slight variation in the range of values at certain pixels can lead to a change in the assigned class label. Therefore, while observing features in the saliency map indicates a more interpretable representation, it does not guarantee an accurate mapping of the actual decision-making process. The advantage of uniquely highlighted regions across multiple classes is that such a saliency map method would resemble a more human-like reasoning process.

\section{Evaluation}

We perform our qualitative and quantitative analysis on two datasets: (1) Wood identification \cite{nieradzik2023automating}: the dataset consists of high-resolution microscopy images for hardwood fiber material. Nine distinct wood species have to be distinguished. (2) DND-Diko-WWWR \cite{wwwr}: This dataset, obtained from unmanned aerial vehicle (UAV) RGB imagery, provides image-level labels for the classification of nutrient deficiencies in winter wheat and winter rye. Classifiers are tasked with distinguishing between seven types of fertilizer treatments.

We trained CNNs for each of these datasets. The ConvNeXt \cite{DBLP:journals/corr/abs-2201-03545} architecture was selected due to its good accuracy on real-world datasets as demonstrated in the research by \cite{fang2023does}. The same paper also shows that many architectures that boast improved accuracy on ImageNet often fail to replicate this performance on real-world datasets. Further substantiating this choice, the study in \cite{nieradzik2023automating} demonstrated ConvNeXt's superior accuracy in wood species identification, surpassing other tested architectures, and performing on par with human experts.

In terms of the fertilizer treatment dataset, we attained approximately 75\% accuracy, representing a robust baseline for our experiments. This ensures that our models are capable at identifying key features, critical for evaluating the attribution maps, across both datasets. Moreover, by submitting our predictions with a slightly stronger model to the fertilizer dataset's competition, we achieved an accuracy $>89$\%, further validating the effectiveness of our models.

\subsection{Consistency and expert annotation}

\subsubsection{Wood identification dataset}

Expert wood anatomists annotated key features within a carefully curated subset of 270 samples from this dataset. These samples were selected for their distinct and easily discernible features. We suspect that the neural network may also base its decision on these features. All experiments were conducted using these 270 samples. Empirical tests confirmed that increases in sample size did not affect the outcomes, thus establishing the statistical significance of the results.

\begin{figure*}[ht]
\begin{subfigure}{.5\textwidth}
  \centering
  \includegraphics[scale=0.5]{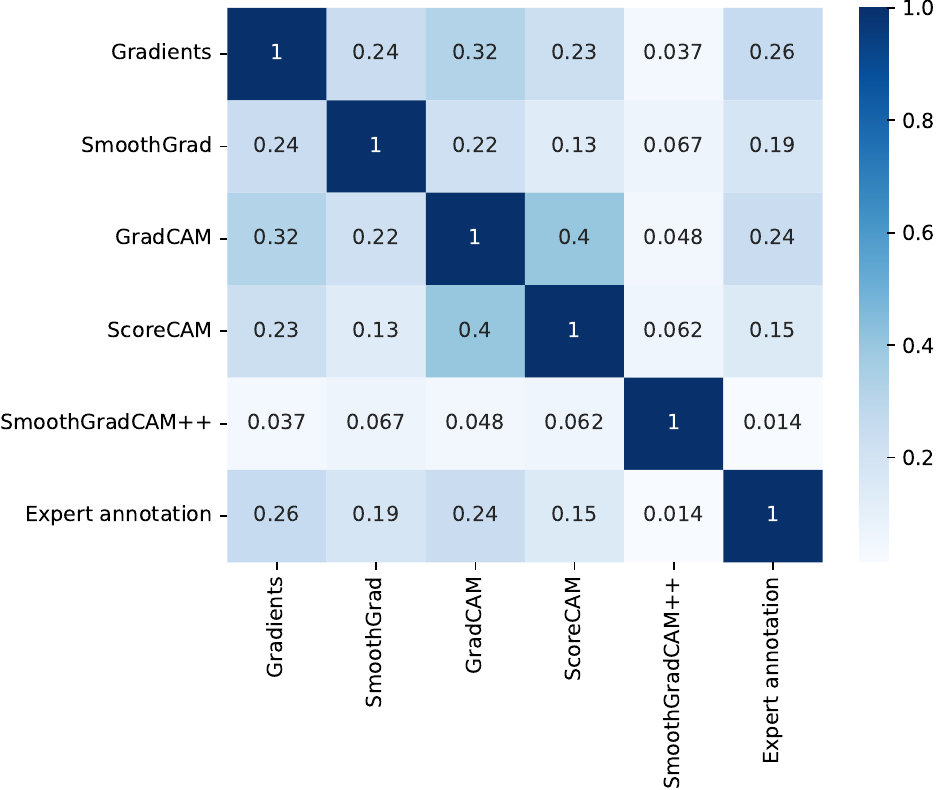}
  \caption{Pearson correlation coefficient (higher is better)}
  \label{fig:a}
\end{subfigure}
\begin{subfigure}{.5\textwidth}
  \centering
  \includegraphics[scale=0.5]{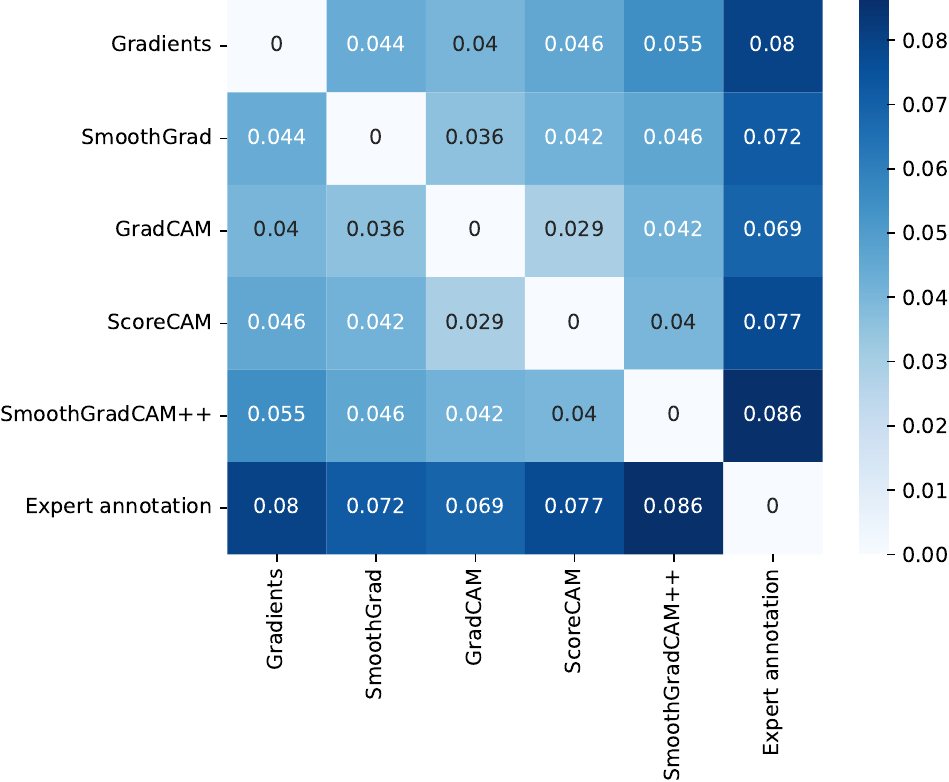}
  \caption{Jensen–Shannon divergence (lower is better)}
  \label{fig:b}
\end{subfigure}
\caption{This plot illustrates the degree of similarity among all attribution maps. The matrices were computed by averaging the individual metric results across all attribution maps in the wood identification dataset. Both similarity measures indicate a weak agreement among the different maps.}
\label{fig:confusionmatrices}
\end{figure*}

For our initial experiment, we examine the consistency among various attribution maps. As demonstrated in \cref{fig:confusionmatrices}, there is only a marginal correlation between different saliency maps. We obtained a $\text{Consistency}_{\text{Corr}}$ of 0.17 and a $\text{Consistency}_{\text{JSD}}$ of 0.05. Both measures similarly rank the consistency across most methods.

Notably, only GradCAM and ScoreCAM demonstrate a reasonably high level of agreement. This can be attributed to the fact that both methods utilize the pre-logit output feature matrix, whereas other methods depend on backpropagation from the output to the input image. However, even in this case, the correlation coefficient shows an agreement barely over 40\%. The visual display of attribution maps for an individual vessel, as depicted in \cref{fig:diffattrmap} on the front page, further underscores this disparity.

This finding is surprising given that the task of wood species identification is well-defined in terms of expert consensus on important features. Hence, the saliency maps' behavior diverges from that of human experts.

Upon considering the expert annotation, we observe that the Gradients method shows the highest similarity with Pearson's correlation coefficient, albeit the agreement falls short of 30\%. Another less apparent issue is the high noise level associated with this method, which complicates the interpretation of results. Conversely, GradCAM and ScoreCAM offer more comprehensible visual explanations, as they are not influenced by backpropagation through multiple distinct layers. In terms of JSD, GradCAM is deemed the most similar to expert annotation, which aligns more closely with expectations from visual inspection.

\subsubsection{Fertilizer treatment dataset}

Similar to the Wood identification dataset, we use around 300 images for performing tests and computing the metrics.

Despite the fertilizer dataset comprising entirely different images, it still exhibits similar inconsistency patterns as witnessed in the previous dataset. With our defined metrics, we measured a $\text{Consistency}_{\text{Corr}}$ of 0.1 and a $\text{Consistency}_{\text{JSD}}$ of 0.06. The correlation coefficient is even lower for this dataset. Both confusion matrices have similar values as the one for the wood identification dataset.

\begin{figure*}[ht]
  \centering
  \includegraphics[scale=0.5]{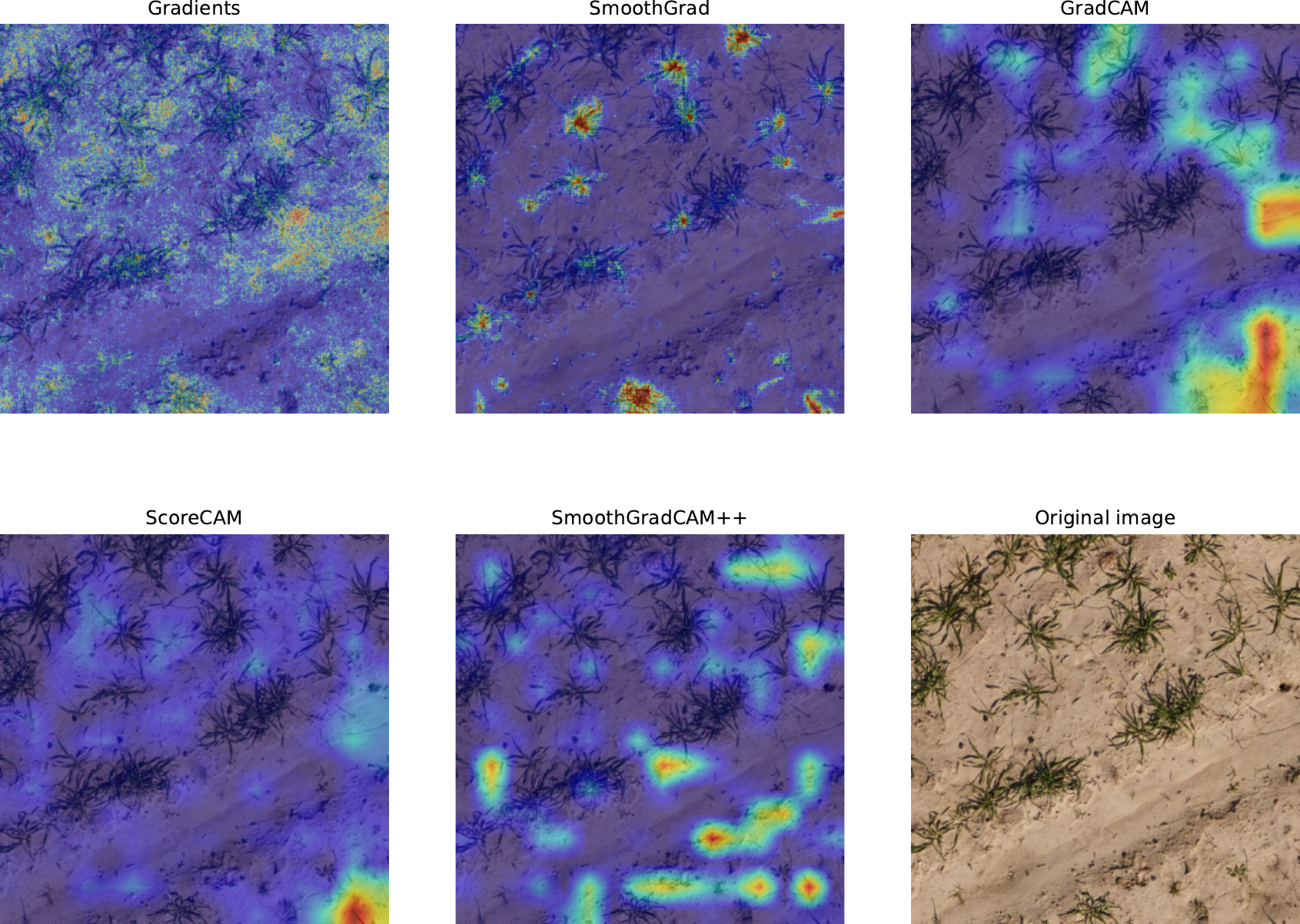}
  
   \caption{Visualization of different attribution maps for the fertilizer dataset of an individual image. Similar to the maps for the wood identification dataset, they exhibit inconsistency in identifying what they consider important.}
   \label{fig:diffattrmap2}
\end{figure*}

As demonstrated visually in \cref{fig:diffattrmap2}, all the attribution maps identify varying regions as significant. In parallel with our earlier observations, we note a substantial amount of noise within the method Gradients.

\subsection{Metrics and feature sharing}

As seen in the previous section, attribution maps lack consistency, underscoring the importance of selecting the most suitable map for the task at hand.

\subsubsection{Wood identification dataset}

We evaluated the maps using two metrics -- "Insertion" and "Deletion", as seen in \cref{tab:delins1}.

\begin{table}[ht]
  \centering
\scalebox{0.9}{\begin{tabular}{llllllllll}
        \toprule
        Attribution Method & Deletion $\downarrow$ & Insertion $\uparrow$ \\ \hline
GradCAM & 0.4622 & $\mathbf{0.9899}$\\
ScoreCAM & 0.5302 & 0.9783\\
SmoothGradCAM++ & 0.5885 & 0.9372\\
\midrule
Gradients & $\mathbf{0.2183}$ & 0.9099\\
SmoothGrad & 0.2741 & 0.9552\\
\midrule
Expert annotation & 0.6868 & 0.8908\\
        \bottomrule
        \end{tabular}}
  \caption{Evaluation of the attribution methods using "Deletion" and "Insertion". The metrics do not provide a conclusive answer on which saliency map is best.}
  \label{tab:delins1}
\end{table}

Interestingly, the metrics suggest differing "best" attribution methods: "Deletion" points to Gradients as superior, while "Insertion" favors GradCAM. Furthermore, neither metric rates the expert annotation highly. This divergence might indicate that the neural network is learning different features from those that human experts would typically recognize.

However, an alternative explanation could be that these metrics themselves may not be fully suitable as evaluation measures. Ideally, we would employ a "metric of a metric" that evaluates the effectiveness of the evaluation metrics themselves. In absence of such a measure, we find ourselves in a situation where "Insertion" and "Deletion" suggest different "best" attribution maps. Hence, visual inspection becomes extremely important when metrics alone cannot conclusively guide the selection of an attribution map.

\begin{figure*}[ht]
\begin{subfigure}{.5\textwidth}
  \centering
  \includegraphics[scale=0.5]{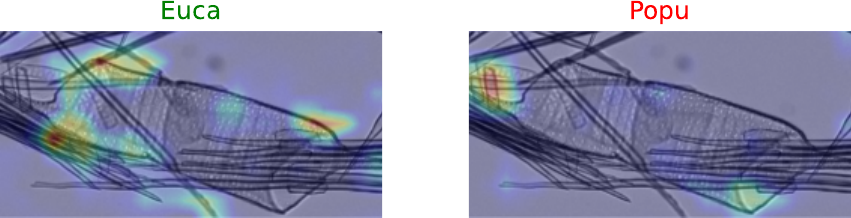}
  \caption{GradCAM}
\end{subfigure}
\begin{subfigure}{.5\textwidth}
  \centering
  \includegraphics[scale=0.5]{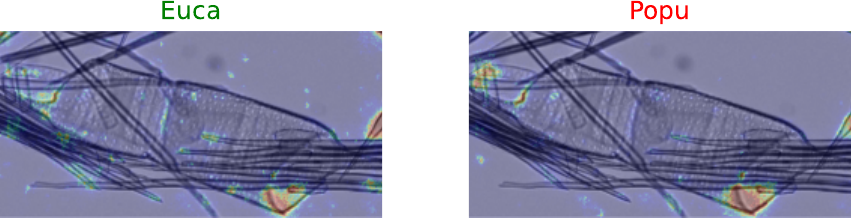}
  \caption{SmoothGrad}
\end{subfigure}
\caption{Attribution maps are intended to visualize the most crucial regions influencing the decision for a specific class in a given model. However, this image comparison reveals that both saliency maps highlight vastly different regions for each class, even though we aim to visualize the same model and classes. SmoothGrad tends to highlight the same regions regardless of the correct class (Euca), whereas GradCAM emphasizes distinct regions. This lack of consistency raises uncertainty about which region the model truly deems most important, making it challenging to identify easily interpretable features for humans.}
\label{fig:diffclass}
\end{figure*}

For this reason, we also explore the visualization for incorrect classes. \Cref{fig:diffclass} illustrates vessel elements from two species: Eucalyptus and Populus. For each image, we provide visualizations for both the correct and incorrect classes. As can be observed from the images, the SmoothGrad saliency map seems to highlight almost identical pixels in both instances. While we cannot definitively determine the correctness of the saliency map's decision-making process without knowledge of the ground truth, this behavior makes it challenging to identify unique features that distinguish between classes. When the same regions are highlighted for two different classes, the interpretation of results becomes more difficult.

From the perspective of a human expert, highlighted regions represent distinctive features that determine the class. In this regard, GradCAM performs more like a human expert. As observed in the figure, the image is partitioned into distinct areas. Although certain regions may be highlighted multiple times, there is a trend of assigning specific regions to a single class, enhancing interpretability.

The behavior of ScoreCAM and SmoothGradCAM++ closely aligns with GradCAM, as all three methods are based on the last feature map. These methods demonstrate a tendency to excel in identifying unique features, mirroring human-like behavior. On the other hand, Gradients exhibits a behavior similar to SmoothGrad in that it assigns almost equal importance to regions across different classes.

\subsubsection{Fertilizer treatment dataset}

We repeated the previous experiment for the fertilizer dataset as can be seen in \cref{tab:delins2}.

\begin{table}[ht]
  \centering
\scalebox{0.9}{\begin{tabular}{lll}
        \toprule
        Attribution Method & Deletion $\downarrow$ & Insertion $\uparrow$ \\ \hline
GradCAM & $\mathbf{0.201}$ & $\mathbf{0.674}$\\
ScoreCAM & 0.274 & 0.573\\
SmoothGradCAM++ & 0.32 & 0.48\\
\midrule
Gradients & 0.312 & 0.451\\
SmoothGrad & 0.289 & 0.546\\
        \bottomrule
        \end{tabular}}
  \caption{Evaluation of the attribution methods using "Deletion" and "Insertion".}
  \label{tab:delins2}
\end{table}

This time the two metrics are more consistent. However, the reliability of these metrics remains questionable, as their consistency varies across different datasets. Additionally, it is important to establish the efficacy of these metrics through rigorous mathematical analysis, ensuring they perform as intended. Currently, the metrics rely primarily on intuitive reasoning and logical arguments, emphasizing the need for further investigation to establish their soundness.

\begin{figure*}[ht]
\begin{subfigure}{.5\textwidth}
  \centering
  \includegraphics[scale=0.42]{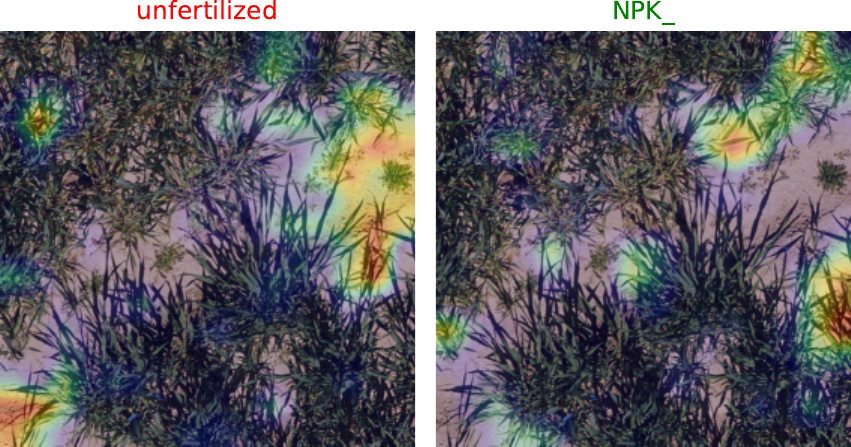}
  \caption{GradCAM}
\end{subfigure}
\begin{subfigure}{.5\textwidth}
  \centering
  \includegraphics[scale=0.42]{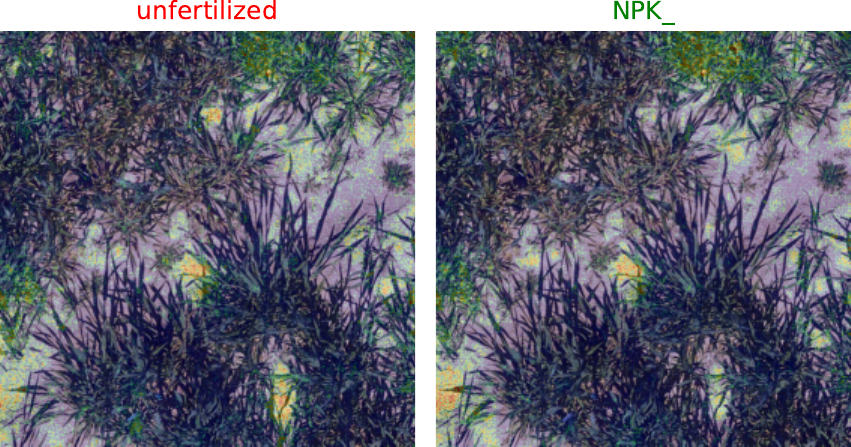}
  \caption{Gradients}
\end{subfigure}
\caption{Visualization of the incorrect (red) and correct class (green) for two saliency maps. The label "NPK\_" correspond to different nutrient statuses: nitrogen, phosphorous, and without potassium. Similar to the wood identification dataset experiment, GradCAM emphasizes distinct regions for each class, while full-backpropagation methods like Gradients tend to focus on the same region for all classes.}
\label{fig:diffclass2}
\end{figure*}

\Cref{fig:diffclass2} showcases the impact of visualizing the saliency map for both incorrect and correct classes of an image. In the case of the incorrect class assumption where the plants were unfertilized, GradCAM predominantly emphasizes the soil. However, when assuming the plants are fertilized (correct class), GradCAM shifts its focus towards the plants. On the other hand, Gradients consistently highlights the plants for both assumptions. This observation suggests that Gradients may place greater importance on the difference in pixel value range between the two classes, whereas GradCAM relies on distinct regions to determine the class. Without knowing the ground truth decision-making process, both approaches are possible.


\section{Discussion and Outlook}

In this study, we conducted a comprehensive evaluation of multiple state-of-the-art attribution maps using real-world datasets from the agriculture and forestry domains. Our analysis unveiled several crucial findings that shed light on the challenges and limitations of these methods.

Firstly, we discovered a significant lack of consistency among attribution maps, both qualitatively and quantitatively. Different methods often highlighted different regions as important, resulting in inconsistent region highlighting and excessive noise in certain approaches. This inconsistency raises concerns about the reliability and robustness of attribution maps for interpreting neural network decisions in agriculture and forestry tasks. We proposed two new metrics for comparing the consistency of attribution maps: Pearson's correlation coefficient and Jensen-Shannon divergence. Both metrics indicated weak agreement among the maps.

Furthermore, when we compared the attribution maps with expert annotations in the wood identification dataset, none of the maps showed high alignment with expert annotations. This suggests that the neural network may learn different features from those identified by human experts, highlighting a disparity in the decision-making process. However, it is also plausible that the maps themselves have limitations in reflecting the true decision-making process of the neural network.

Another important observation was the excessive feature sharing across distinct classes in certain attribution maps. This behavior can make it challenging to interpret the results and uncover features that are easily interpretable by humans. The ability to identify unique features for each class is crucial for gaining a better understanding of the decision-making process of neural networks and providing meaningful explanations.

Our evaluation of attribution maps using the metrics "Insertion" and "Deletion" showed certain limitations. The parameters can have a significant influence on the results and the metrics are in general not consistent across datasets. It is worth emphasizing that many CAMs \cite{https://doi.org/10.48550/arxiv.2006.14255, wang2020scorecam, DBLP:journals/corr/abs-2010-03023} have asserted their superiority based on the "Insertion" and "Deletion" metrics. Given our findings, which highlight the unreliability of these metrics, a  question arises: Do these CAMs genuinely demonstrate improvements compared to GradCAM?

It is necessary to prove that the evaluation metrics actually work. Otherwise, selecting the most suitable attribution map for a specific task can only be based on visual inspection and comparison with human annotation.

In conclusion, our research highlights the effectiveness and reliability challenges of attribution maps for wood species identification and fertilizer treatment classification in the agriculture and forestry domains. The lack of consistency among attribution maps and the disparity between the maps and expert annotations underscore the need for further research and development in this field. Future work should focus on developing improved attribution methods that address the limitations identified in this study. Additionally, there is a necessity to find metrics that can objectively evaluate the attribution maps, as current metrics such as "Insertion" and "Deletion" may lead to conflicting results. By addressing these challenges, we can enhance the interpretability and trustworthiness of neural networks in critical applications within agriculture and forestry.

\bibliographystyle{unsrtnat}
\bibliography{references}

\end{document}